\documentclass[letterpaper, 10 pt, conference]{ieeeconf}  % Comment this line out if you need a4paper

\IEEEoverridecommandlockouts                              % This command is only needed if 
\usepackage{graphics} % for pdf, bitmapped graphics files
\usepackage{epsfig} % for postscript graphics files
\usepackage{times} % assumes new font selection scheme installed
\usepackage{amsmath} % assumes amsmath package installed
\usepackage{amssymb}  % assumes amsmath package installed

\usepackage[utf8]{inputenc} % allow utf-8 input
\usepackage[T1]{fontenc}    % use 8-bit T1 fonts

\usepackage{pdfx}
\usepackage{xcolor}
\usepackage{color}

\makeatletter
\let\NAT@parse\undefined
\makeatother
\usepackage{hyperref}
\hypersetup{
    colorlinks=true,
    linkcolor=orange,
    filecolor=magenta,      
    urlcolor=orange,
    citecolor=orange,
}
\usepackage{url}            % simple URL typesetting
\usepackage{booktabs}       % professional-quality tables
\usepackage{amsfonts}       % blackboard math symbols
\usepackage{nicefrac}       % compact symbols for 1/2, etc.
\usepackage{microtype}      % microtypography

\usepackage{cite}

\usepackage{caption}

\usepackage{graphicx}
\usepackage{subfigure}
\usepackage{booktabs} % for professional tables
\usepackage{multirow}
\usepackage{pifont}
\usepackage[noend]{algpseudocode}
\usepackage{setspace}
\usepackage{comment}
\usepackage[ruled,linesnumbered]{algorithm2e}
\usepackage{blindtext}
\usepackage{verbatim}
\newcommand{\WSCL}{{WeaSCL}}

\usepackage{pdfx}

\title{\LARGE \bf Weakly Supervised Correspondence Learning}

\author{Zihan Wang$^{*{1}}$, Zhangjie Cao$^{*{2}}$, Yilun Hao$^{3}$ and Dorsa Sadigh$^{4}$% <-this % stops a space
\thanks{$^{1}${\tt\small wangzih@stanford.edu} $^{2}${\tt\small caozj@cs.stanford.edu}, ${3}${\tt\small yilunhao@stanford.edu} and $^{3}${\tt\small dorsa@cs.stanford.edu}. The authors are with the Department of Computer Science, Stanford University, Stanford, CA 94305, USA}
\thanks{* means Equal Contribution. Author ordering determined by coin flip over a Google Hangout.}
 }

\begin{document}

\maketitle
\thispagestyle{empty}
\pagestyle{empty}

%%%%%%%%%%%%%%%%%%%%%%%%%%%%%%%%%%%%%%%%%%%%%%%%%%%%%%%%%%%%%%%%%%%%%%%%%%%%%%%%
\begin{abstract}
Correspondence learning is a fundamental problem in robotics, which aims to learn a mapping between state, action pairs of agents of different dynamics or embodiments. 
However, current correspondence learning methods either leverage strictly paired data---which are often difficult to collect---or learn in an unsupervised fashion from unpaired data using regularization techniques such as cycle-consistency---which suffer from severe misalignment issues.
We propose a weakly supervised correspondence learning approach that trades off between strong supervision over strictly paired data and unsupervised learning with a regularizer over unpaired data. 
Our idea is to leverage two types of weak supervision: i) temporal ordering of states and actions to reduce the compounding error, and ii) paired abstractions, instead of paired data, to alleviate the misalignment problem and learn a more accurate correspondence.
The two types of weak supervision are easy to access in real-world applications, which simultaneously reduces the high cost of annotating strictly paired data and improves the quality of the learned correspondence. 
%Our experimental results in Mujoco simulation, a simulated robot, and a real robot environment show that our method substantially outperforms prior works in various correspondence learning settings including cross-morphology, cross-physics, and cross-modality. 
We show the videos of the experiments on our \href{https://sites.google.com/stanford.edu/weakly-supervised-correspond}{website}.
\end{abstract}

% keywords: Machine Learning for Robot Control Transfer Learning Representation Learning

\section{Introduction}
Humans are born with the ability to develop new skills by mimicking the behavior of others who may have different embodiments~\cite{nehaniv2002correspondence}. For example, prior cognitive science work suggest that 1- or 2-year-old children can infer the intentions of adults and re-enact their behavior with their own body even with a large difference in body structures~\cite{nielsen200912,meltzoff1995understanding}. We refer to the ability to infer the mapping between the state, action pairs of agents with different dynamics or embodiment as \emph{correspondence learning}. Correspondence learning is essential in robotics where we have limited data and would like to learn from demonstrations from other agents.

To learn the correspondence between agents, several prior works leverage paired trajectories to learn invariant representations across agents~\cite{gupta2017learning,sermanet2018time,jha2018disentangling,zhang2021learning}, where the representation only preserves the information that is relevant to the downstream tasks. However, collecting and annotating paired trajectories require experts with substantial domain knowledge and is usually expensive to access at large scale. 

Due to the difficulties of collecting paired data, several works propose learning the correspondence between environments as a translation map between the agents using unpaired trajectories~\cite{zhu2017unpaired,bansal2018recycle}. The key insight of these works is adopting a regularization term over the translation model, where cycle-consistency is the most commonly used regularization~\cite{smith2019avid,hoffman2018cycada,james2019sim,zhang2020learning}. However, with no supervision, the quality of the learned correspondence model is usually not as good as models learned with strong supervision over paired data~\cite{zhang2020weakly,shukla2019extremely}. 

In this paper, we propose \emph{Weakly Supervised Correspondence Learning} (\WSCL) to find a trade-off between strong supervision of strictly paired data and regularization over unpaired data. \textit{Our key insight is to leverage weak supervision that is useful for learning correspondence and also is easy to access in real-world applications.} We propose two types of weak supervision: i) temporal ordering in states and actions, and ii) paired abstractions over data. 

The \emph{temporal ordering}, which originates from the nature of sequential decisions, indicates the temporal dependency of the consecutive states and actions. Leveraging temporal dependency as a measure of weak supervision enables us to avoid compounding errors of translation maps that can be accumulated over long horizons.

\begin{figure}[t]
\centering
\includegraphics[width=0.5\textwidth]{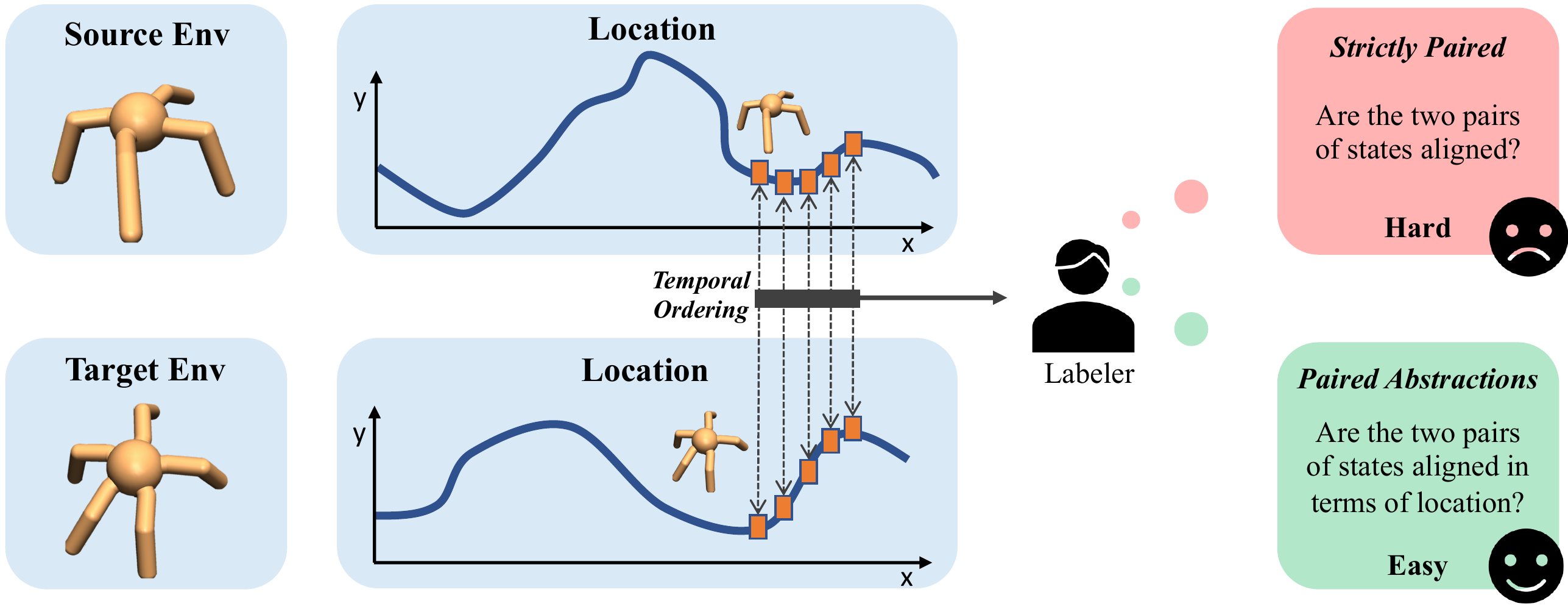}
\caption{\small{An example of the paired abstractions. Given two trajectories of a four- and five-legged ant robots, it is difficult to decide whether two full states that include joint angles of each agent are aligned, while it is easy to align simpler abstractions such as spatial location. 
}}
\vspace{-15pt}
\label{fig:front}
\end{figure}

We define \emph{paired abstractions} by a similarity metric over some abstraction of states or state-action pairs of the agents. For example, the location of a mobile robot, the pose of an end-effector, or the confidence of a behavior can potentially be suitable abstractions over data. When learning correspondence between two agents, one can consider a pair of these abstractions as opposed to paired data.  
The paired abstractions are easier to obtain and annotate than strictly paired data, as annotators would have an easier time comparing similarity over simpler abstractions.
For example, in Fig.~\ref{fig:front}, it would be difficult to align the full states including the joint angles of the trajectories of a four- and five-legged Ant. On the other hand, it is much easier and more informative to decide if an abstraction of the state, e.g., the location of the Ant agents on the 2D plane are aligned. 
In our work, we collect such paired abstractions and learn a similarity function over this data. We then incorporate this similarity function in the loss function imposing a constraint on the translation maps.
In summary, the contributions of this paper are:
\begin{itemize}
    \item We propose a weakly supervised correspondence learning approach to address the shortcomings of using strictly paired data or using unpaired data with regularization. 
    \item Our approach utilizes weak supervision (temporal ordering over states and paired abstractions over data) to learn correspondence. This weak supervision enforces multi-step dynamics cycle-consistency over a sequence of states and actions and also imposes a similarity function learned from paired abstractions as a constraint in correspondence learning.
    \item Our empirical results on cross-morphology, cross-physics, and cross-modality correspondence learning tasks in Mujoco, simulated robot, and real robot environments show that \WSCL\ achieves much higher performance compared to prior methods.
\end{itemize}

\section{Related Works}

\noindent\textbf{Learning Invariant Representations.}
To learn the correspondence across agents, one line of works learn an invariant representation of states and actions, which remove any dependencies on unrelated information for the downstream task and only preserve task-specific information~\cite{sadeghi2016cad2rl,tobin2017domain,peng2018sim,gupta2017learning,sermanet2018time,yan2020learning}. 
Domain randomization methods learn generalizable domain invariant representations by augmenting the current domain, but they require the variation of the applied domains to be covered by the augmentation~\cite{tobin2017domain,pinto2017asymmetric,peng2018sim,sadeghi2016cad2rl,andrychowicz2020learning,ramos2019bayessim,zakharov2019deceptionnet,wu2019learning,chen2020robust}. 
This assumption is restrictive and requires the full domain information to design effective augmentations. Other works remove this assumption and learn invariant representations from paired trajectories~\cite{gupta2017learning,sermanet2018time,liu2018imitation,yan2020learning}. However, supervision over paired trajectories require domain expertise, which is expensive or even impossible to collect~\cite{taylor2009transfer}. Instead of such strong supervision, our approach uses weak supervision to learn the correspondence, which is easier to annotate.

\noindent\textbf{Learning Translation Maps.}
Due to the challenges of collecting paired data, approaches that use unpaired data are proposed to learn a translation map between the agents' trajectories~\cite{taylor2007transfer,ammar2015unsupervised,tzeng2020adapting,joshi2018cross,kim2020domain,smith2019avid,8794310}. Most of the works on learning translation maps are proposed in the visual domains. Cycle-consistency was proposed to address image translation across different domains and it achieves promising results~\cite{zhu2017unpaired}. Many follow up works improve the stability of the training and the quality of the translated images~\cite{zhou2016learning,zhu2017unpaired,liu2018unified,hoffman2018cycada,bansal2018recycle,bousmalis2018using,james2019sim,smith2019avid,rao2020rl}. Recent works propose utilizing weakly aligned images to learn the translation~\cite{tzeng2020adapting}. Going beyond visual observations, Ammar et al. use unsupervised manifold alignment to find the correspondence between states across domains from demonstrations but they rely on hand-designed features, which restricts generalization~\cite{ammar2015unsupervised}. 
Kim et al. propose to imitate demonstrations by building correspondence between the agents but assume the MDPs are `alignable' with respect to a definition of MDP reduction~\cite{kim2020domain}. 

Recently, dynamic cycle-consistency (DCC) is proposed to learn a translation map over the states and actions across domains~\cite{zhang2020learning}. DCC is not restricted to the visual domains and is proven to be applicable to different physics, modalities, and morphologies. 
%The following work builds a transfer imitation learning method based on DCC~\cite{zhang2021policy}. 
Though achieving the state-of-the-art performance with unpaired trajectories, DCC still does not perform as well as methods with strong supervision. Our approach is closely related to DCC, but also imposes weak supervision over DCC to learn a more accurate translation map without the need for highly costly annotations. 

 \noindent\textbf{Learning with Insufficient Annotations.} 
% Currently, the significant performance of deep learning algorithms highly relies on a large amount of labeled data. However, for particular applications, strong supervision that directly provides the annotation of the task to solve can be difficult to attain. Thus, weakly supervised learning is proposed to leverage weak supervision that provides imprecise or inexact but easy-to-access labels~\cite{zhou2018brief}. In the computer vision domain, weak supervision is widely adopted for tasks where strong supervision is difficult to annotate. For example, in semantic segmentation and object detection, per-pixel labels and per-object bounding boxes are difficult to annotate, and object labels or the whole image label~\cite{huang2018weakly,araslanov2020single,bilen2016weakly} are used as weak supervision. In video action localization, the actual action boundary requires watching each frame to annotate and the sequence of action labels~\cite{huang2016connectionist} or the whole video labels are used as supervision~\cite{paul2018w}. 
For particular tasks, the exact annotations of the task are difficult to obtain, which results in different learning frameworks to deal with limited annotations. Semi-supervised learning aims to learn from little labeled data and large-scale unlabeled data. For correspondence learning, the small slice of data can be annotated by keyframes extraction and segmentation~\cite{sakoe1978dynamic,rohrbach2012database}. However, such accurate annotations are sometimes impossible to provide even with expert knowledge. Thus, weakly supervised learning is proposed to leverage weak supervision that provides imprecise or inexact but easy-to-access labels~\cite{zhou2018brief}. Weakly supervised learning has been used in robotics and control tasks such as goal-orientated reinforcement learning~\cite{lee2020weakly} and goal-directed navigation~\cite{ma2019towards}, which substantially reduces the exploration space. However, in correspondence learning, prior works often only consider strong supervision, i.e., using strictly paired trajectories or they only rely on regularization along with unpaired data. In this work, we focus on leveraging weakly supervised learning in correspondence learning.

\section{Correspondence Learning:\\ Problem and Background}
In this section, we introduce the problem of correspondence learning and provide some background on dynamic cycle-consistency first introduced by~\cite{zhang2020learning}.

\noindent \textbf{Correspondence Learning.}
We focus on learning correspondence between two agents. However, we note that one can extend this to multiple agents by building correspondence between pairs of agents.
We model each agent as a deterministic Markov Decision Process (MDP): $\mathcal{M}^1=(\mathcal{S}^1, \mathcal{A}^1, \mathcal{T}^1, \mathcal{R}^1, p^1_0, \gamma)$ and $\mathcal{M}^2=(\mathcal{S}^2, \mathcal{A}^2, \mathcal{T}^2, \mathcal{R}^2, p^2_0, \gamma)$. 
Similar to~\cite{zhang2020learning}, we define a correspondence from $\mathcal{M}^1$ to $\mathcal{M}^2$ as follows:
Let $\Phi:\mathcal{S}^1 \rightarrow \mathcal{S}^2$ be a state map, and $H^1:\mathcal{S}^1 \times \mathcal{A}^1 \rightarrow \mathcal{A}^2$ and $H^2:\mathcal{S}^2 \times \mathcal{A}^2 \rightarrow \mathcal{A}^1$ be two action maps, where the state map and the action maps satisfy the following requirements: $\forall s^1\in \mathcal{S}^1$, if $s^2=\Phi(s^1)$, then $\forall a^1\in \mathcal{A}^1, \Phi(\mathcal{T}^1(s^1, a^1))=\mathcal{T}^2(s^2,H^1(s^1,a^1))$ and $\forall a^2\in \mathcal{A}^2, \Phi(\mathcal{T}^1(s^1,H^2(s^2,a^2)))=\mathcal{T}^2(s^2, a^2)$. Intuitively, the requirements mean that the successor states of the two aligned states should be aligned if taking aligned actions. 

Using this correspondence definition, we are now ready to introduce our problem statement.
We assume access to three pieces of information: a set of trajectories (sequence of state, action pairs) $\Xi^1=\{\xi^1\}$ for $M^1$, a set of trajectories $\Xi^2=\{\xi^2\}$ for $M^2$, and one or multiple sets of paired abstractions over the states or over the state-action pairs. Specifically, we have $K^s$ sets of paired abstractions over states: $Y_1^s, Y_2^s,\dots,Y_{K^s}^s$ and $K^a$ sets of paired abstractions over state-action pairs. Each $Y_{k}^s$ is a set of pairs of states and similarity labels over abstractions of states: $Y_{k}^s=\{(s^1, s^2,v^s)\}$, where $v^s\in [0,1]$ reflects the similarity of one choice of abstraction, e.g., the pose of an end-effector, over the state $s^1$ and $s^2$. Note that the data tuples $(s^1, s^2,v^s)$ are given by annotators, where the annotators decide which abstraction to take and how to annotate similarity. Our algorithm does not have access to the choice of abstraction and similarity but aims to learn a similarity function $\Phi^\text{weak}_k:\mathcal{S}^1 \times \mathcal{S}^2 \rightarrow [0,1] $ mapping the raw pairs of states to a similarity value based on the given data tuples.
Similarly each $Y_k^a=\{((s^1, a^1), (s^2, a^2), v^a)\}$ and $v^a\in [0,1]$ reflects the similarity of a choice of abstraction over $(s^1, a^1)$ and $(s^2, a^2)$, and we aim to learn a similarity function $H^\text{weak}_k:\mathcal{S}^1 \times \mathcal{A}^1 \times \mathcal{S}^2 \times \mathcal{A}^2 \rightarrow [0,1]$ mapping the raw pairs of state-action pairs to the similarity value.
Our goal in correspondence learning is to learn the state map $\Phi$ and the action maps $H^1$ and $H^2$ with $\Xi^1$, $\Xi^2$, and the similarity functions learned from the paired abstraction data $Y_1^s,\dots,Y_{K^s}^s$ and $Y_1^a,\dots,Y_{K^a}^a$.

We emphasize that the paired abstractions only consider a loose alignment between the states and actions of the two MDPs. %For example, as in Fig.~\ref{fig:front}, using paired abstractions, we only consider pairing of locations of the two Ant agents as opposed to strict alignment of the full states including the joint angles or velocities. 
Such loose pairing of the states---pairing of abstractions over states---simply can be assessed by visual observations, and collecting such data along with annotations is much easier, and can serve as a cheap supervision.

\noindent \textbf{Background on Dynamics Cycle-Consistency.}
% To address our correspondence learning problem, we would like to first introduce some background on dynamic cycle-consistency (DCC)~\cite{zhang2020learning}.
%To differentiate the notation of ground-truth functions and learned functions, we use $\hat{\cdot}$ to indicate the learned functions, e.g., we use $\hat{\Phi}$ to indicate the learned state map. 
Dynamic Cycle-Consistency (DCC)~\cite{zhang2020learning} first uses adversarial learning to ensure that the states mapped by $\Phi$ fall into the domain of $\mathcal{M}^2$. Specifically, one can learn $\Phi$ with a discriminator $D^s$ by the following adversarial objective:
\begin{equation}\label{eqn:gans}
\begin{small}
\begin{aligned}
    &\min\limits_{\Phi}\max\limits_{D^s}\mathcal{L}^s_\text{adv}(\Phi, D^s)=\\
    &\mathbb{E}_{s^2 \sim \Xi^2}[D^s(s^2)] + \mathbb{E}_{s^1 \sim \Xi^1}[1-D^s(\Phi(s^1))].
\end{aligned}
\end{small}
\end{equation}
In addition, DCC ensures that the actions mapped by $H^1$ and $H^2$ also match the actions in the domain of $\mathcal{M}^2$ and $\mathcal{M}^1$ using discriminators $D^{a^1}$ and $D^{a^2}$ respectively:
\begin{equation}\label{eqn:gana}
\begin{small}
\begin{aligned}
    &\min\limits_{H^1,H^2}\max\limits_{D^{a^1},D^{a^2}}\mathcal{L}^a_\text{adv}(H^1,H^2,D^{a^1},D^{a^2})= \\
 & \quad\mathbb{E}_{a^2 \sim \Xi^2}[D^{a^2}(a^2)] + \mathbb{E}_{(s^1,a^1) \sim \Xi^1}[1-D^{a^2}(H^1(s_1,a^1))]\\
    &+\mathbb{E}_{a^1 \sim \Xi^1}[D^{a^1}(a^1)] + \mathbb{E}_{(s^2,a^2) \sim \Xi^2}[1-D^{a^1}(H^2(s_2,a^2))].
\end{aligned}
\end{small}
\end{equation}
Finally, one can add a domain cycle-consistency objective on the state-action maps $H^1$ and $H^2$:
\begin{equation}\label{eqn:domcon}
\begin{small}
\begin{aligned}
    &\min\limits_{H^1,H^2}\mathcal{L}_\text{dom\_con}(H^1,H^2)=\\
    &\quad\mathbb{E}_{(s^1,a^1)\in \Xi^1}\left[\lVert H^2\left(\Phi(s^1), H^1(s^1,a^1)\right)-a^1\rVert\right] \\
    &+ \mathbb{E}_{(s^2,a^2)\in \Xi^2}\left[\lVert H^1\left(\Phi(s^2), H^2(s^2,a^2)\right)-a^2\rVert\right].
    \end{aligned}
\end{small}
\end{equation}
This equation ensures that the two action maps are consistent with each other and the translated action should be able to be translated back.

The adversarial training as proposed so far suffers from the mode collapse problem~\cite{goodfellow2014generative}, where multiple states for one agent can potentially be mapped to one state in the other. In addition, the domain cycle-consistency cannot solve the problem when the two maps $H^1$ and $H^2$ make consistent mistakes. For example, we can map $(s^1,a^1)$ to an incorrect action, e.g., $\bar{a}^2$, by $H^1$ and map it back to $a^1$ by $H^2$. Here, both maps make mistakes but the domain consistency is still preserved. To address this issue, DCC introduces the dynamics cycle-consistency objective:
\begin{equation*}
\begin{small}
\begin{aligned}
    &\min\limits_{\Phi,H^1}\mathcal{L}_\text{dyn\_con}(\Phi,H^1)=\\
    &\quad \mathbb{E}_{(s^1_t,a^1_t,s^1_{t+1})\sim \Xi^1}\left[\left\lVert\Phi(s^1_{t+1}) - \mathcal{T}^2\left( \Phi(s^1_t), H^1(s^1_t,a^1_t) \right)\right\rVert\right].
    \end{aligned}
\end{small}
\end{equation*}
Here, the transition function $\mathcal{T}^2$ for $\mathcal{M}^2$ is not always known and can be non-differentiable. So one can empirically learn a transition function $\hat{\mathcal{T}}^2$ using the following objective:
\begin{equation*}\label{eqn:forward}
\begin{small}
    \min\limits_{\hat{\mathcal{T}}^2}\mathcal{L}_\text{forward}(\hat{\mathcal{T}}^2)=\mathbb{E}_{(s^2_t,a^2_t,s^2_{t+1})\sim \Xi^2}\left[\left\lVert s^2_{t+1}-\hat{\mathcal{T}}^2(s^2_t,a^2_t) \right\rVert\right].
\end{small}
\end{equation*}
Combining all the losses introduced so far, the final optimization objective is:
\begin{equation*}
\begin{aligned}
    \mathcal{L}_\text{DCC}=&\lambda_0 \mathcal{L}_\text{dyn\_con}(\Phi,H^1) + \lambda_1\mathcal{L}_\text{dom\_con}(H^1,H^2)\\
    &+\lambda_2\mathcal{L}^a_\text{adv}(H^1,H^2,D^{a^1},D^{a^2}) +\lambda_3 \mathcal{L}^s_\text{adv}(\Phi, D^s),
\end{aligned}
\end{equation*}
where $\lambda_0$, $\lambda_1$, $\lambda_2$ and $\lambda_3$ are hyperparameters trading off between the different losses. DCC firstly trains the forward dynamics $\hat{\mathcal{T}}^2$ and then trains the translation model with $\mathcal{L}_\text{DCC}$.

\noindent \textbf{Limitations of DCC.}\label{sec:limit}
Here, we discuss two core shortcomings of DCC---compounding error and misalignment---which can lead to errors in the translation model.

\begin{figure}[ht]
    \centering
    \subfigure[Compounding Error]{\includegraphics[width=.235\textwidth]{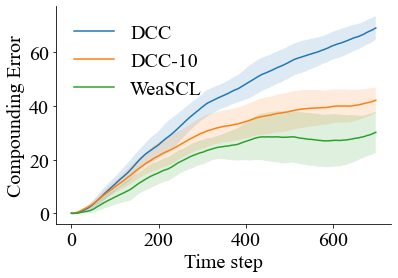}\label{fig:compounding_error}}
    \subfigure[Different Horizon]{\includegraphics[width=.235\textwidth]{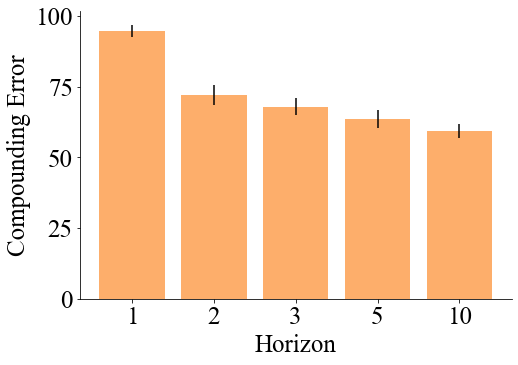}\label{fig:horizon}}
    \caption{ \small{(a) The translation error at each time step. (b) The compounding error with respect to different final horizons.
    % (a) The difference of states at different future steps derived by ground-truth actions and by translated actions; (b) The difference of x-axis location (major term in the reward) for the HalfCheetah environment with different horizons to perform dynamics cycle-consistency.
    }}
    \label{fig:}
    \vspace{-10pt}
\end{figure}

The compounding error problem refers to the fact that the single step errors from the state and actions maps can accumulate over a sequence.
We empirically demonstrate the existence of compounding errors by selecting a segment of a trajectory with horizon $T$: $\xi^1=\{s^1_0,a^1_0,\dots,s^1_T\}$ in $\Xi^1$. We use two methods to derive the translated state at time step $T$: (1) $s^2_T=\Phi(s^1_T)$; (2) $\hat{s}^2_T=\mathcal{T}^2\left(\cdots\mathcal{T}^2\left(\Phi(s^1_0), H_1(s^1_0, a^1_0)\right),\dots, H_1(s^1_T, a^1_T)\right)$. 
The second method continuously uses the translated action to generate the next state to follow the transition process in $\xi^1$.
%Comparing the states reached by the first and second approach, we can empirically check whether there exists compounding errors if we consecutively use the translated actions. 
We experiment in the Mujoco HalfCheetah environment to build a correspondence between the two-legged and three-legged robots. As shown in Fig.~\ref{fig:compounding_error}, the distance of $s^2_T$ and $\hat{s}^2_T$ for DCC gets larger over time, which suggests the existence of compounding errors in the action maps. 
Our hypothesis is that this is due to the fact that dynamics cycle-consistency is only ensured for one time step and leads to a small error in that step but cannot bound the error over a long horizon.

Dynamic Cycle-Consistency still suffers from misalignment issues. 
For example, assume we are given two trajectories $\xi^1_A$ and $\xi^1_B$ for the agent following $\mathcal{M}^1$ and two trajectories $\xi^2_A$ and $\xi^2_B$ for the agent following $\mathcal{M}^2$, where the four trajectories have the same number of time steps. 
Let's assume the ground-truth translation should translate $\xi^1_A$ to $\xi^2_A$ and $\xi^1_B$ to $\xi^2_B$. 
However, if one only enforces dynamics cycle-consistency, it is possible to learn a map that translates the states and actions at each step from $\xi^1_A$ to $\xi^2_A$ and from $\xi^1_B$ to $\xi^2_B$, or translates from $\xi^1_A$ to $\xi^2_B$ and from $\xi^1_B$ to $\xi^2_A$, where both maps have zero errors in terms of dynamics cycle-consistency. 
So the misalignment issue can occur without strong supervision of paired data. However, strictly paired data is often difficult to collect, and we thus aim for some intermediate supervision such as learning similarities between paired abstractions over states, which are much easier to annotate.

\section{Weakly Supervised Correspondence Learning}
We propose weakly supervised correspondence learning (\WSCL) to address the above issues with two weak supervision: temporal ordering and paired abstraction data.

\noindent \textbf{Multi-Step Dynamics Cycle-Consistency.}
As we discussed in Sec.~\ref{sec:limit}, even a small error for the state map and the action maps at each step will cause a large deviation in a long horizon because DCC only enforces one-step consistency and the error can accumulate across time steps given no constraint. To address this problem, we use the weak supervision of consecutive states and actions to enforce the dynamics cycle-consistency over multiple steps. 
The new loss can be formulated as follows:
\begin{equation}\label{eqn:dyncon}
\begin{small}
\begin{aligned}
    &\min\limits_{\Phi,H^1}\mathcal{L}_\text{m\_dyn\_con}(\Phi,H^1)= \mathbb{E}_{(s^1_t,a^1_t,s^1_{t+1},\cdots,s^1_{t+T})\sim \Xi^1}\sum\limits_{\tau=1}^{T}\\
    &\quad\left[\left\lVert\Phi(s^1_{t+\tau}) - \hat{\mathcal{T}}^2\left(\cdots\hat{\mathcal{T}}^2\left( \Phi(s^1_t), \hat{a}^2_t\right)\cdots \hat{a}^2_{t+\tau-1}\right) \right\rVert\right],
    \end{aligned}
\end{small}
\end{equation}
where $\hat{a}^2_t = H_1(s^1_t,a^1_t)$ is the translated action at time $t$ and $T$ is the final horizon to enforce dynamics cycle-consistency. With this new loss, as shown in Fig.~\ref{fig:compounding_error}, with final horizon $10$, the compounding error is substantially reduced.

Now we should consider how long to enforce the dynamics cycle-consistency. %If the horizon is small, the compounding error problem could still exist. If the horizon is too large, computing the loss at each step can lead to high computation costs.
We conduct an experiment on the performance of translation with respect to the final horizon in the HalfCheetah environment.
We create two agents $\mathcal{M}^1$ with three legs and $\mathcal{M}^2$ with two legs. We translate the states of $\mathcal{M}^1$ to $\mathcal{M}^2$ with $\Phi$ and take the optimal action based on the optimal policy of $\mathcal{M}^2$. We then translate the action back to $\mathcal{M}^1$ with $H^2$.
In Fig.~\ref{fig:horizon}, we observe that the performance of translation increases with a longer horizon at first but saturates from horizon $5$ onwards. 
% The observation indicates that we can treat the final horizon as a hyperparameter and tune it by gradually increasing the horizon until when the performance saturates.

\noindent \textbf{Learning Correspondence by Weak Supervision.}
To address the misalignment issue, we adopt weak supervision from paired abstractions, where a similarity metric is defined on the abstractions (e.g. the location, end-effector pose, or confidence score) % over a state or state-action pair as opposed to the full information.
The key difference between strictly paired data and paired abstractions is that strictly paired data need to comprehensively assess all the aspects of the two states or state-action pairs, which is difficult to collect. On the other hand, paired abstractions only consider similarities over an abstraction of the state, which are thus easier to annotate.

\begin{figure}
    \centering
    \includegraphics[width=.35\textwidth]{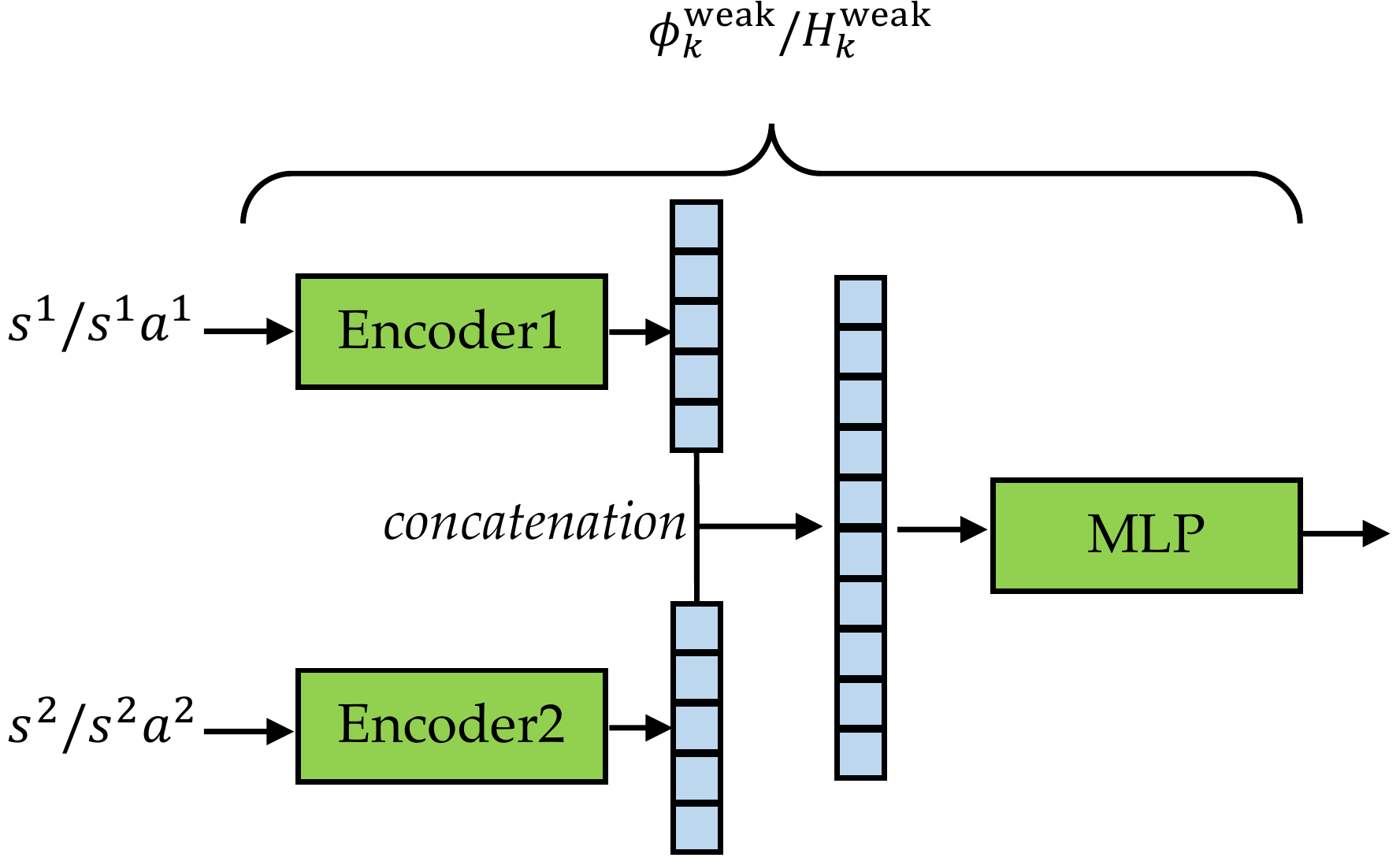}
    \caption{The architecture of the similarity function. The states or state-action pairs from the two agents are first mapped by their individual encoders to a shared hidden space. The hidden features are concatenated and mapped to the similarity value with a multi-layer perceptron.}
    \vspace{-15pt}
    \label{fig:sim_fig}
\end{figure}

We first learn a similarity function from each set of paired abstraction data, which is modelled as a neural network with a pair of states or state-action pairs as input and outputs a similarity value in $[0,1]$. The architecture is shown in Fig.~\ref{fig:sim_fig}. We first map the input states from both agents to the same hidden space by their individual encoder and concatenate the two hidden features. Then we use a fully-connected network to map the concatenated feature to the scalar similarity value.
The losses for all the similarity functions are 
\begin{equation}\label{eqn:weak}
\begin{small}
    \begin{aligned}
    & \min\limits_{\Phi^\text{weak}_{k}} \mathcal{L}^s_k(\Phi^\text{weak}_{k}) = \mathbb{E}_{(s^1,s^2,v^s) \sim Y_k^s} \ell({\Phi^\text{weak}_{k}(s^1,s^2)}, v^s) \\
    &  \min\limits_{H^\text{weak}_{k}} \mathcal{L}^a_k(H^\text{weak}_{k}) =\\
    &\quad\quad\quad\mathbb{E}_{((s^1,a^1),(s^2,a^2),v^s) \sim Y_k^s} \ell({H^\text{weak}_{k}(s^1,a^2,s^2,a^2)}, v^s),\\
    \end{aligned}
\end{small}
\end{equation}
where $\ell$ takes the binary cross entropy loss to minimize the difference between the predicted and the ground-truth similarity. Then, we impose the learned similarity function as a constraint on the state map and the action maps:
\begin{equation}\label{eqn:weak_min}
\begin{small}
    \begin{aligned}
&\min\limits_\Phi \mathcal{L}^\text{weak}_s(\Phi) =  \sum\limits_{k=1}^{K^s}\mathbb{E}_{s^1\in \Xi^1}\left[-\Phi^\text{weak}_k(s^1,\Phi(s^1))\right]\\
   &\min\limits_{H^1}\mathcal{L}^\text{weak}_a(H^1) =\\  &\quad\sum\limits_{k=1}^{K^a}\mathbb{E}_{(s^1,a^1)\in \Xi^1}\left[-H^\text{weak}_k(s^1, a^1,\Phi(s^1),H^1(s^1, a^1))\right].\\
    \end{aligned}
\end{small}
\end{equation}
We minimize the negative similarity to ensure the states and the translated states are similar as well as the state-action pairs and the translated state-action pairs stay similar. 
With the above constraint, the misalignment of the learned translation model will be substantially reduced. Also, as shown in Fig.~\ref{fig:compounding_error}, paired abstractions can reduce the compounding error by reducing the translation error at each step. 

\noindent\textbf{Overall Loss and Algorithm.} Integrating all the losses, we derive the final learning objective of our model as follows:
\begin{equation}\label{eqn:all}
\begin{small}
\begin{aligned}
    \mathcal{L}_\text{all}=&\lambda_0\mathcal{L}_\text{m\_dyn\_con}(\Phi,H^1) + \lambda_1\mathcal{L}_\text{dom\_con}(H^1,H^2)\\
    &+\lambda_2\mathcal{L}^a_\text{adv}(H^1,H^2,D_{a^1},D_{a^2}) +\lambda_3 \mathcal{L}^s_\text{adv}(\Phi, D_s)\\
    &+\lambda_4(\mathcal{L}^\text{weak}_s(\Phi)+\mathcal{L}^\text{weak}_a(H^1))\\
    %&+\sum\limits_{k=1}^{K^s}\mathcal{L}^s_k(\Phi^\text{weak}_{k})+\sum\limits_{k=1}^{K^a}\mathcal{L}^a_k(H^\text{weak}_{k}),
\end{aligned}
\end{small}
\end{equation}
where $\lambda_4$ is the trade-off parameter for the weakly supervised loss. 
Jointly optimizing all the loss functions in Eqn.~\eqref{eqn:all} can cause unstable training ~\cite{zhang2021learning}.
 Thus, we first learn the forward model $\hat{\mathcal{T}}^2$ and the similarity functions $\Phi^\text{weak}_1-\Phi^\text{weak}_{K^s}$ and $H^\text{weak}_1-H^\text{weak}_{K^a}$. After converging, we fix their parameters. %We do not fine-tune these parameters during correspondence learning since the parameters are already learned to differentiate similar or dissimilar pairs and we can use them to decide whether the mapped state/state-action pair is similar to the original state/state-action pair. 
 Then, we iteratively train the networks related to the state map: $\Phi$ and $D^s$, and the networks related to the action maps: $H^1$, $H^2$, $D^{a^1}$ and $D^{a^2}$. When we train $\Phi$ and $D^s$, we fix the parameters of $H^1$, $H^2$, $D^{a^1}$, and $D^{a^2}$, and vice versa. Such an iterative training paradigm avoids the state map and the action maps converging to unstable solutions.
% coordinating to find a 'shortcut' solution.
When training $\Phi$ and $D^s$ or $H^1,H^2$, and $D^{a^1},D^{a^2}$, we follow the training paradigm of adversarial networks~\cite{goodfellow2014generative}.

\section{Experiments}
In our experiments, we aim to demonstrate the efficacy of \WSCL\ in different correspondence learning settings including cross-morphology, cross-physics, and cross-modality, and demonstrate that \WSCL\ works well with different types of paired abstractions in different environments. 

We use \textbf{\WSCL-$T$} to refer to our approach, where $T$ corresponds to the final horizon at which we enforce dynamics cycle-consistency. We compare \WSCL-$T$ with baseline methods: \textbf{DCC}~\cite{zhang2021learning} and \textbf{CC}, which removes the dynamics cycle-consistency in DCC, and several variants of \WSCL: \textbf{DCC-$T$} and \textbf{\WSCL-1}, where DCC-$T$ only adopts multi-step dynamics cycle-consistency without using paired abstractions while \WSCL-1 uses the paired abstractions but only uses single-step dynamics cycle-consistency.

% We show additional experimental details and more visualization results on our \href{https://sites.google.com/stanford.edu/weakly-supervised-correspond}{website}.

\subsection{Cross-Morphology}

\begin{table}[ht]
    \vspace{-5pt}
    \centering
        \caption{Morphology parameters and dimension of state and action spaces in the HalfCheetah, Swimmer and Ant.}
    \label{tab:morphology_setup}
    \resizebox{.49\textwidth}{!}{
    \begin{tabular}{c|ccc|ccc}
    \toprule
\multirow{2}{50pt}{Environment} & \multicolumn{3}{c|}{Agent $\mathcal{M}^2$} & \multicolumn{3}{c}{Agent $\mathcal{M}^1$}\\
& Morphology & State & Action & Morphology & State & Action \\
    \midrule
    HalfCheetah & 2 legs &  18 &  6 & 3 legs &  24 &  9  \\
    Swimmer & 3 links &  10 &  2 & 4 links &  12 &  3 \\
    Ant & 4 legs &  113 &  8 & 5 legs &  135 &  10 \\
    \bottomrule
    \end{tabular}}
    \vspace{-5pt}
\end{table}
\noindent\textbf{Mujoco Environments.} We conduct our experiments in Mujoco HalfCheetah, Swimmer, and Ant environments under a \textbf{cross-morphology} setting, where we create different agents by varying the morphology. The morphology and the dimension of state space and action space are shown in Table~\ref{tab:morphology_setup}. The goal of this task is to learn and evaluate a translation model to leverage the optimal policy for the agent $\mathcal{M}^1$ to make decisions in the environment of agent $\mathcal{M}^2$. We measure the similarity of states using the x-axis location as the abstraction of the state. Since both state spaces and action spaces are different, we train both the state map $\Phi$ and action maps $H^1$ and $H^2$. The number of similarity pairs used for all three environments are 1,000 each.  %We evaluate the performance of the translation model by the performance of the translated policy in $\mathcal{M}^2$.

The results are shown in Table~\ref{tab:result_morphology}. For both DCC and our methods, using a horizon of $5$ for dynamics cycle-consistency achieves a much better performance than a horizon of $1$, which demonstrates the efficacy of multi-step dynamics cycle-consistency. \WSCL-5 and \WSCL-1 outperform DCC-5 and DCC-1 respectively, which demonstrates the efficacy of paired abstractions.

\begin{table}[ht]
%\vspace{-10pt}
    \centering
    \addtolength{\tabcolsep}{-3pt}
    \caption{The performance of the translated policy under different morphologies in Mujoco environments.}
    \label{tab:result_morphology}
    \resizebox{.4\textwidth}{!}{
    \begin{tabular}{c|cccc}
    \toprule
         Method & HalfCheetah & Swimmer & Ant\\
         \midrule
        CC & -104.39$\pm$92.72 & 30.00$\pm$2.19 & 297.52$\pm$87.48  \\
        DCC-1 & 658.66$\pm$23.13 & 53.40$\pm$11.39 & 447.50$\pm$470.19  \\
        \midrule
        DCC-2 & 1005.52$\pm$44.12 & 64.92$\pm$5.43 & 669.94$\pm$72.54 \\
        DCC-3 & 1166.90$\pm$ 50.67 & 71.70$\pm$3.53 & 762.43$\pm$1.92 \\
        DCC-5 & 1250.55$\pm$ 51.66 & 65.19$\pm$ 2.16 & 928.22$\pm$ 1.96 \\
        DCC-10 & 1249.15$\pm$434.78 & 52.18$\pm$3.61 & 942.03$\pm$ 2.61 \\
        \WSCL-1 & 1284.61$\pm$109.47 & 69.59$\pm$13.88 & 969.28$\pm$1.03 \\
        \midrule
        \WSCL-5 & \textbf{1455.08}$\pm$63.59 & \textbf{86.14}$\pm$2.46 & \textbf{971.08}$\pm$2.10 \\
        \midrule
        Oracle & 4380.75$\pm$97.30 & 126.19$\pm$2.42 & 991.56$\pm$1.98  \\
         \bottomrule
    \end{tabular}}
    \vspace{-5pt}
\end{table}

\begin{figure}[ht]
\vspace{-5pt}
    \centering
    \subfigure{\includegraphics[width=.45\textwidth]{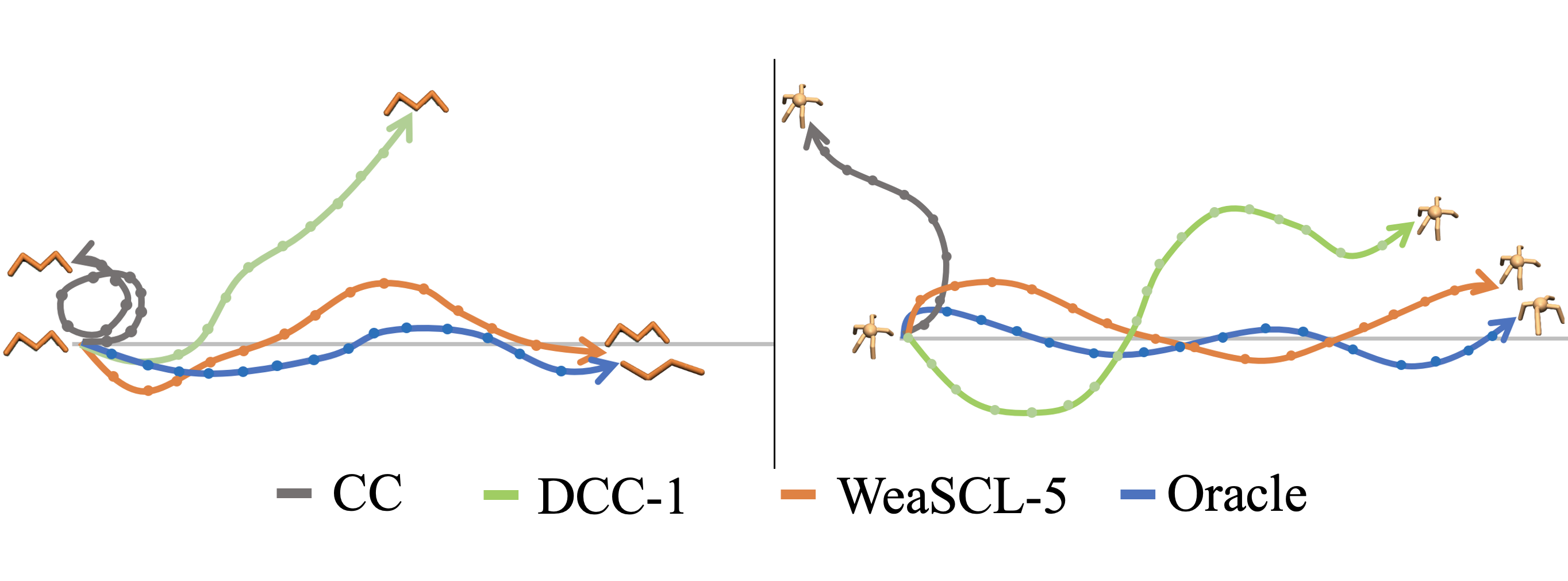}\label{fig:trajectories}} 
    \vspace{-20pt}
    \caption{Sample trajectories for 4-link swimmer (left) and 5-legged ant (right). The grey line is the positive x-axis, which direction the robot is supposed to move toward. The oracle is only available in $\mathcal{M}^2$ (3-link swimmer and 4-legged ant).
    }
    \label{fig:swimmer_ant_trajectories}
    \vspace{-10pt}
\end{figure}

\noindent\textbf{Simulated Robots.}
As shown in Fig.~\ref{fig:simulated_robot}, we create two dynamics in the simulated Panda Robot: the original 7-DoF robot arm, and a 5-DoF arm that fixes the third and forth joints of the 7-DoF arm (shown by red crosses). We define the paired abstractions based on the end-effector position in the state (green arrows) or the joint force in the action (purple arrows). We test two settings of paired abstractions: (1) only using the end-effector position ($Y^s$); (2) using both the end-effector position and the joint force ($Y^s$ and $Y^a$). Our goal is to translate the policy from 5-DoF to 7-DoF.

We show our results in Table~\ref{tab:result_simulated}. We observe that \WSCL-5 outperforms the baselines, DCC-1 and CC. \WSCL-5 also outperforms the variants: \WSCL-1 and DCC-5, which demonstrates the efficacy of both kinds of weak supervisions. We also note that \WSCL-5 with $Y^s$ and $Y^a$ outperforms \WSCL-5 with $Y^s$, which demonstrates that \WSCL can handle similarities over multiple abstractions elegantly and having access to similarities over multiple types of abstractions improves the performance.

\begin{minipage}{0.47\textwidth}
\vspace{5pt}
\begin{minipage}{.48\textwidth}
    \centering
    \includegraphics[width=\textwidth]{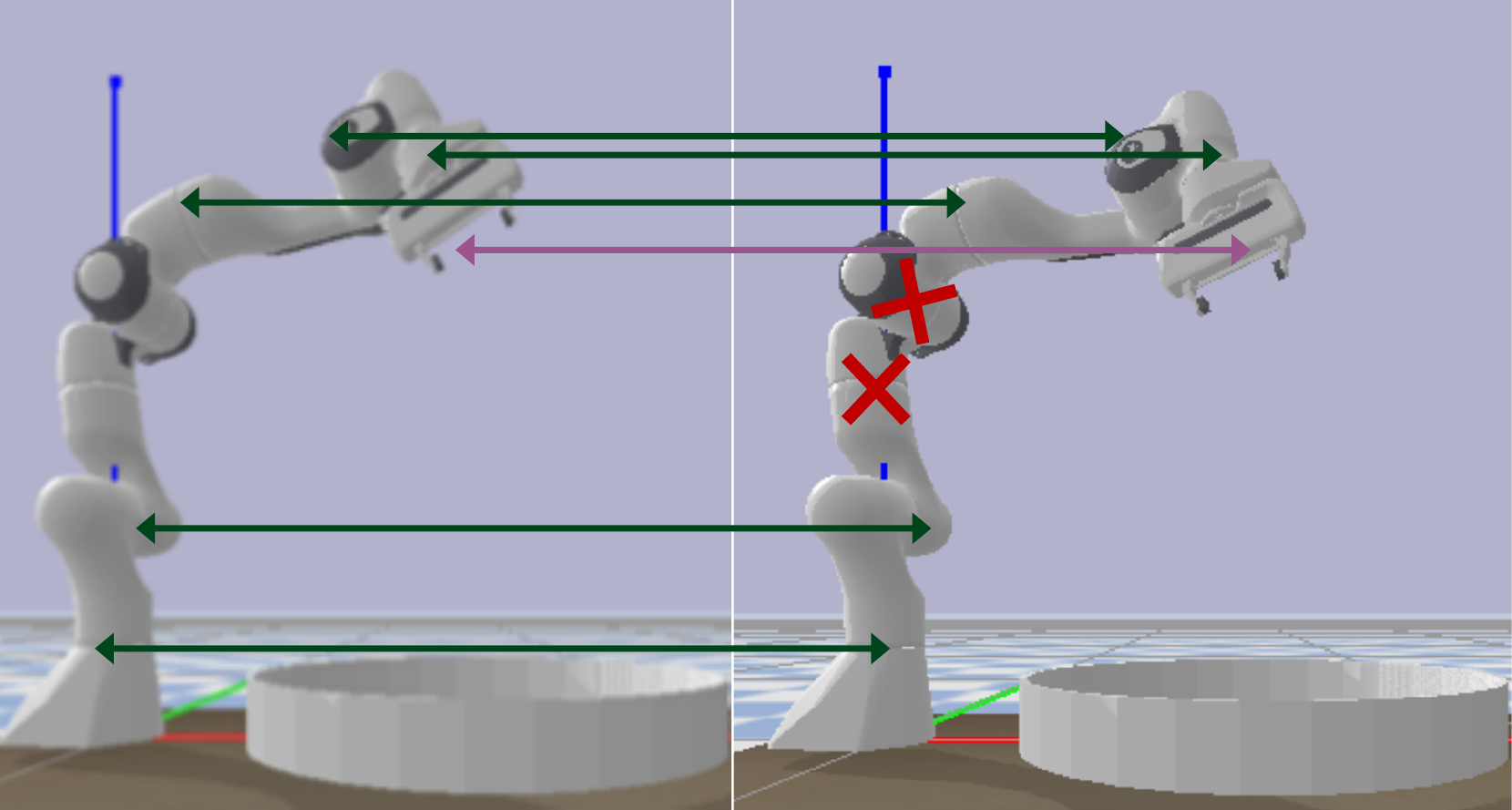}
    \captionof{figure}{\small{Demonstrating the two robot arms with different degrees of freedom and the paired abstractions of end-effector positions and joint forces.}}
    \label{fig:simulated_robot}
\end{minipage}
\begin{minipage}{.5\textwidth}
\addtolength{\tabcolsep}{-3pt}
    \centering
    \resizebox{\textwidth}{!}{
    \begin{tabular}{c|cccc}
    \toprule
        CC & -315.09$\pm$115.74   \\
        DCC-1 & -255.33$\pm$160.19  \\
        \midrule
        DCC-5 & -233.47$\pm$103.19  \\
        \WSCL-1 ($Y^s$) & -225.28$\pm$100.39 \\
        \WSCL-1 ($Y^s$ and $Y^a$) & -219.88$\pm$121.11\\
        \midrule 
        \WSCL-5 ($Y^s$) & -78.43$\pm$22.06  \\
        \WSCL-5 ($Y^s$ and $Y^a$) & \textbf{-73.07} $\pm$42.19 \\
        \midrule
        Oracle & -20.68$\pm$21.30   \\
         \bottomrule
    \end{tabular}}
            \captionof{table}{\small{The performance of the translated policy under different morphologies in the simulated robot environment.}}
            \label{tab:result_simulated}
\end{minipage}
\vspace{5pt}
\end{minipage}

\subsection{Cross-Physics}

We conduct the experiments in Mujoco Hopper and Walker2d environments under a \textbf{cross-physics} setting, where we create different agents by varying the physical factors in the environment. We vary the gravitational constant in the Hopper environment and vary the friction of feet in the Walker2d environment. The exact value of the gravitational constant and the friction of the agent $\mathcal{M}^1$ and $\mathcal{M}^2$ are in Table~\ref{tab:physics_setup}. Note that only changing the physical parameters does not change the state and action spaces but changes the transition. %where taking the same action at a state will transition to potentially different states, so we only learn the action maps to align the agents. 
Our goal is to translate a policy across environments with different physical parameters.

\begin{table}[ht]
    \centering
        \caption{Physical parameters in the Hopper and Walker2d.}
    \label{tab:physics_setup}
    \resizebox{.46\textwidth}{!}{
    \begin{tabular}{c|c|cc}
    \toprule
    \multirow{2}{100pt}{Environment} & \multirow{2}{50pt}{Agent $\mathcal{M}^2$} & \multicolumn{2}{|c}{Agent $\mathcal{M}^1$}\\
    & & Setup1 & Setup2\\
    \midrule
    Hopper (Gravitational Constant) & 9.8 & 0.5 & 5.0 \\
    Walker2d (Friction) & 0.9 & 9.9 & 19.9 \\
    \bottomrule
    \end{tabular}}
    \vspace{-5pt}
\end{table}

We use confidence as the abstraction to define similarity, where confidence lies in $[0,1]$ indicating how good a state, action pair is with respect to the reward function. For example, if a state, action pair always appears in optimal trajectories, we regard it as optimal and assign confidence $1$ to it. Then for all the state-action pairs in all trajectories, we randomly sample $1000$ pairs of state-action pairs with varying similarity as the dataset to learn the similarity function.

Here, we need trajectories with different confidence values for $\Xi^1$ and $\Xi^2$. For each environment and physical parameter, we train $7$ policies with different rewards, which range from the random policy to the optimal policy. We then collect $10$ trajectories from each policy as the trajectory set. We compute the reward for each trajectory and normalize the reward into $[0,1]$ by min-max normalization, where the normalized reward is used as the confidence for each trajectory. For each state-action pair in a trajectory, we use the trajectory confidence value as the confidence used for abstraction.

\begin{table}[ht]
    \centering
    \addtolength{\tabcolsep}{-3pt}
    \caption{The performance of the transferred or translated policy under different physics.}
    \label{tab:result_physics}
    \resizebox{.5\textwidth}{!}{
    \begin{tabular}{c|cccc}
    \toprule
         Method & Gravity $0.5$ & Gravity $5.0$ & Friction $9.9$ & Friction $19.9$\\
         \midrule
        Direct & 269.59$\pm$2.45 & 335.41$\pm$7.89 & 290.84$\pm$10.12 & 280.49$\pm$20.54 \\
        CC & 61.19$\pm$39.91 & 83.26$\pm$155.79 & 178.93$\pm$219.81 & 236.15$\pm$72.39\\
        DR & 295.64$\pm$4.87 & 376.31$\pm$9.41 & 297.32$\pm$9.42 & 310.18$\pm$22.24 \\
        DCC-1 & 26.48$\pm$45.17 & 6.03$\pm$4.32 & 305.28$\pm$7.01 & 375.22$\pm$101.77  \\
        \midrule
        DCC-2 & 271.59$\pm$50.02 & 190.08$\pm$186.22 & 369.07$\pm$48.11 & 588.70$\pm$201.02 \\
        DCC-3 & 234.46$\pm$217.32 & 229.69$\pm$243.62 & 302.15$\pm$7.11 & 540.31$\pm$143.63 \\
        DCC-4 & 256.91$\pm$55.10 & 195.03$\pm$148.78 & 307.50$\pm$3.04 & 799.97$\pm$138.63 \\
        DCC-5 & 276.73 $\pm$120.39 & 231.76$\pm$161.27 & 305.11$\pm$4.62 & 598.56$\pm$219.53 \\
        \WSCL-1 & 208.14$\pm$189.65 & 143.72$\pm$180.01 & 321.04$\pm$14.40 & 587.73$\pm$117.49 \\
        \midrule
        \WSCL-2 & \textbf{325.80}$\pm$57.06 & 279.33$\pm$107.24 & \textbf{499.52}$\pm$48.99 & \textbf{1052.62}$\pm$224.62 \\
        \WSCL-3 & 137.49$\pm$132.33 & \textbf{387.12} $\pm$186.80 & 301.30$\pm$4.18 & 693.94$\pm$245.45 \\
        \WSCL-4 & 129.05$\pm$84.62 & 283.38 $\pm$184.15 & 308.94$\pm$2.39 & 674.47$\pm$122.79 \\
        \WSCL-5 & 130.09$\pm$75.48 & 272.69$\pm$91.31 & 306.67$\pm$6.30 & 550.24$\pm$202.03 \\
        \midrule
        Oracle & 1952.99$\pm$32.41 & 3060.55$\pm$21.72 & 3604.38$\pm$52.59 & 1632.18$\pm$22.86 \\
         \bottomrule
    \end{tabular}}
    \vspace{-5pt}
\end{table}

%We train the similarity function for $10$ epochs, the forward dynamics model for $20$ epochs, and the translation model for $30$ epochs. We use Adam optimizer with a learning rate $0.001$ for all the model trainings. We use trade-offs parameters $\lambda^0=20,\lambda_1=1,\lambda_2=1,\lambda_3=30$.

We show the results of our method and baselines in Table~\ref{tab:result_physics}. For DCC and our method, we report the results of using the dynamics cycle-consistency for $1-5$ steps, since the performance does not increase or even decrease for more than $5$ steps. We observe that our method with a proper number of steps for dynamics cycle-consistency achieves the best reward in all the tasks. Note that in most of the tasks, only two steps of dynamics cycle-consistency are sufficient to achieve the best performance, which demonstrates that the proposed approach is computationally efficient.

\subsection{Cross-Modality}

\begin{figure}[ht]
\vspace{-10pt}
    \centering
    \subfigure{\includegraphics[width=.38\textwidth]{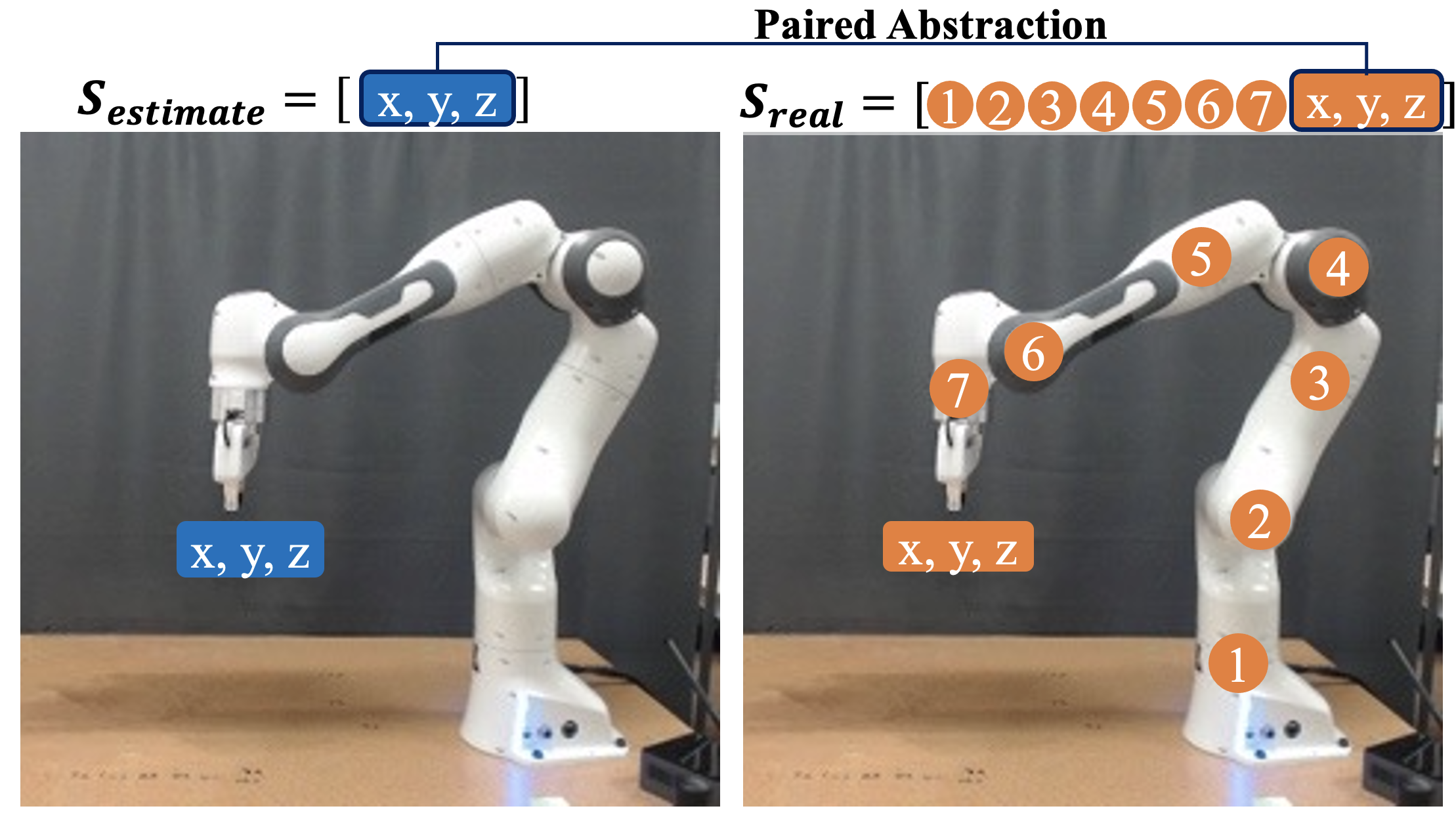}\label{fig:sim_real_robot}} \\
    \vspace{-5pt}
    \subfigure{\includegraphics[width=.28\textwidth]{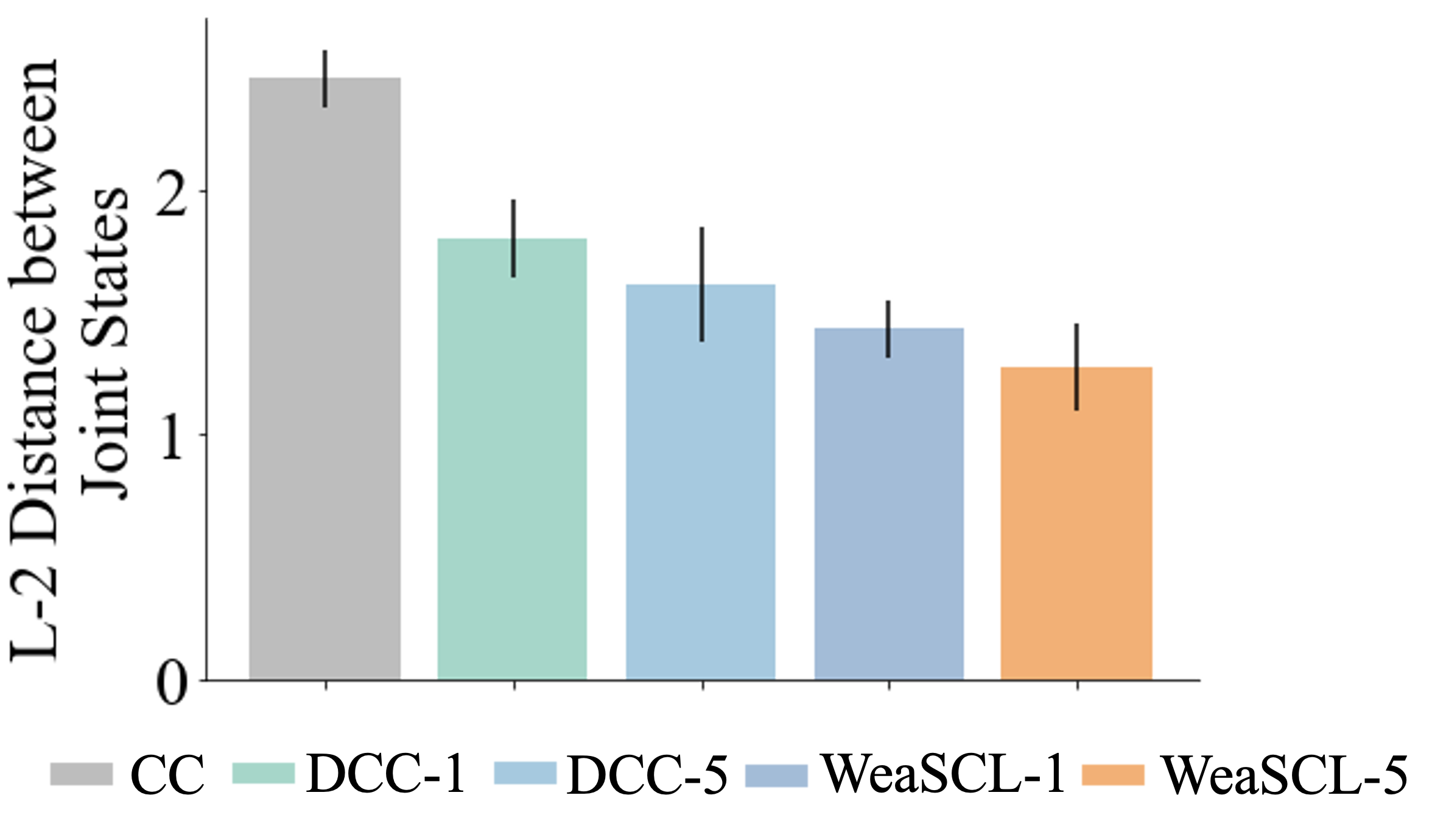}\label{fig:real_robot_exp}}
    \vspace{-10pt}
    \caption{ Top: An illustration of the real robot arm environments. Bottom: The norm of joint differences on the robot.
    }
    \label{fig:cross_modality}
    \vspace{-7pt}
\end{figure}

We also conduct experiments on the real robot under a \textbf{cross-modality} setting, where we translate across the visual observations and the joint states of a real Franka Panda robot arm. 
Our goal is to predict the state of the robot (joint configurations) from the visual observation of the robot. Our abstraction here is the end-effector pose of the robot in these two domains (ground-truth state and visual observations) and we collect $100$ similarity pairs to learn the similarity function..
% The goal of the experiment is to estimate the joint configurations by given the real robot images. 
% We measure the L-2 distance between the estimate joint state and the ground truth joint state. 
% The abstraction is defined as the end effector position of the robot estimated from the visual observation or from the joint states.
Note that the actions are the same and we just need to learn the state map $\Phi$, which takes the RGB images as inputs and outputs the joint state of the robot. %The visual observations are captured by an external RGB camera with a third-person point of view. We collect random trajectories of visual observations and joint configurations (not paired) on the robot through teleoperation to train the state map $\Phi$.

As shown in Figure~\ref{fig:cross_modality}, \WSCL-5 achieves the lowest estimation error compared to baselines CC and DCC-1 and also the variants DCC-5 and \WSCL-1, which demonstrates the efficacy of our approach in real robot applications.

% \begin{table}[ht]
%     \centering
%     \addtolength{\tabcolsep}{-3pt}
%     \caption{The Norm of Joint Differences on Real Panda Robot.}
%     \label{tab:cross_modality}
%     \resizebox{.42\textwidth}{!}{
%     \begin{tabular}{c|cc|ccc}
%     \toprule
%          Env & CC &DCC-1 &DCC-5 & WSCL-Weakly & WSCL-5 \\
%          \midrule
%         Panda-Real & 2.461$\pm$0.119 & 1.807$\pm$0.160 & 1.618$\pm$0.235 & 1.433$\pm$0.114& \textbf{1.277}$\pm$0.179  \\

%          \bottomrule
%     \end{tabular}}
% \end{table}

\section{Conclusion}
\noindent \textbf{Summary.} We propose a weakly supervised correspondence learning approach (\WSCL) that leverages weak supervision in the form of temporal ordering and paired abstraction data. This eases the need for expensive paired data, and enables more accurate correspondence learning. Experiment results show that \WSCL\ outperforms the state-of-the-art correspondence learning methods based on unpaired data.

\noindent \textbf{Limitations and Future Work.} 
Though we leverage the easy-to-access weak supervision to improve correspondence learning, this type of supervision still requires domain knowledge or human experts to annotate. 
% One potential future direction is to learn from unlabeled or unpaired data. 
In the future, we also plan to automatically detect the abstraction needed for weak supervision and reduce the size of the required annotation.
 
% \section{Acknowledgements}
\section{Acknowledgements}
We would like to thank FLI grant RFP2-000, NSF Awards 1849952 and 1941722, and ONR for their support.

{\small
\bibliography{main.bib}
\bibliographystyle{IEEEtran}
}

\end{document}

\maketitle

\section{Correspondence Definition for Stochastic MDPs}

\section{Algorithm}
Jointly optimizing all the networks causes the training process to be unstable and makes networks to take 'shortcut' solutions as mentioned in~\cite{zhang2021learning}. Thus, in Algorithm~\ref{alg:algo1}, we first learn the forward model $\hat{\mathcal{T}}^2$ and the weak correspondent functions $\Phi^\text{weak}_1-\Phi^\text{weak}_{K^s}$ and $H^\text{weak}_1-H^\text{weak}_{K^a}$. In Algorithm~\ref{alg:algo2}, after the training of these networks converges, we fix their parameters. Then, we alternatively training the networks related to the state map: $\Phi$ and $D_s$, and the networks related to the action maps: $H_1$, $H_2$, $D_{a^1}$ and $D_{a^2}$. When we train $\Phi$ and $D_s$, we fix the parameters of $H_1$, $H_2$, $D_{a^1}$ and $D_{a^2}$ and vice versa. Such alternative training paradigm avoids the state map and the action maps coordinating to find a 'shortcut' solution. On Line 3-14, following training paradigm of adversarial networks~\cite{goodfellow2014generative}, we train $\Phi$ and $D_s$ alternatively. On Line 15-28, we train the action maps $H_1$ and $H_2$, and the discriminators $D_{a^1}$ and $D_{a^2}$ alternatively.

\begin{algorithm}[h]
\KwIn{The demonstration set $\Xi^1$ of the agent $\mathcal{M}^1$ and $\Xi^2$ of the agent $\mathcal{M}^2$. The loosely paired data for states $Y^s_1,\cdots,Y^s_{K^s}$ and for actions $Y^a_1,\cdots,Y^a_{K^a}$. The maximal horizon $T$ to ensure dynamics cycle-consistency.}
  Initialize $\hat{\mathcal{T}}^2$, $\Phi^\text{weak}_1-\Phi^\text{weak}_{K^s}$ and $H^\text{weak}_1-H^\text{weak}_{K^a}$\;
  \For{$k=1\rightarrow K^s$}{
  \While {not converging}{
 Sample a batch of loosely paired states $\{(s^1, s^2, v^s)\}$ from $Y_k^s$\;
 Train $\Phi^\text{weak}_k$ with $\{(s^1, s^2, v^s)\}$ according to the objective in Eqn. (8)\;}}%~\eqref{eqn:weak}
 \For{$k=1\rightarrow K^a$}{
 \While {not converging}{
 Sample a batch of loosely paired states $\{((s^1,a^1), (s^2,a^2), v^a)\}$ from $Y_k^a$\;
 Train $H^\text{weak}_k$ with $\{((s^1,a^1), (s^2,a^2), v^a)\}$ according to the objective in Eqn. (8)\;}%~\eqref{eqn:weak}
 }
 \While {not converging}{
 Sample a batch of state, action and next state tuples $\{s^2_t,a^2_t,s^2_{t+1}\}$ from $\Xi^2$\;
  Train $\hat{\mathcal{T}}^2$ with $\{s^2_t,a^2_t,s^2_{t+1}\}$ according to the objective in Eqn. (5)\;%~\eqref{eqn:forward}
 }
  \KwOut{The learned forward model $\hat{\mathcal{T}}^2$, and the learned weak correspondent functions $\Phi^\text{weak}_1-\Phi^\text{weak}_{K^s}$ and $H^\text{weak}_1-H^\text{weak}_{K^a}$.}
  \caption{Algorithm for learning the forward model and the weak correspondent functions}\label{alg:algo1}
\end{algorithm}

\begin{algorithm}[h]
\KwIn{The demonstration set $\Xi^1$ of the agent $\mathcal{M}^1$ and $\Xi^2$ of the agent $\mathcal{M}^2$. The maximal horizon $T$ to ensure dynamics cycle-consistency. The learned $\hat{\mathcal{T}}^2$, $\Phi^\text{weak}_1-\Phi^\text{weak}_{K^s}$ and $H^\text{weak}_1-H^\text{weak}_{K^a}$.}
  Initialize $\Phi$, $H_1$, $H_2$, $D_s$, $D_{a^1}$, $D_{a^2}$\;
  
  \While {not converging}{
  Fix the parameters of $H_1$, $H_2$, $D_{a^1}$ and $D_{a^2}$\;
   \For{$i=1\rightarrow N^s$}{
  Sample a batch of state-action pairs $\{s^1\}$ from $\Xi^1$ and $\{s^2\}$ from $\Xi^2$\;
  Compute $\mathcal{L}^s_\text{adv}$ with $\{s^1\}$ and $\{s^2\}$ according to Eqn. (1)\;%~\eqref{eqn:gans}
  Sample a batch of segments $\{s^1_t,a^1_t,\cdots,s^1_{t+T}\}$ from $\Xi^1$\;
  Compute $\mathcal{L}_\text{m\_dyn\_con}$ with $\{s^1_t,a^1_t,\cdots,s^1_{t+T}\}$ according to Eqn. (7)\;%~\eqref{eqn:dyncon}
  Sample a batch of states $\{s^1\}$ and state-action pairs $\{(s^1,a^1)\}$ from $\Xi^1$\;
  Compute $\mathcal{L}^\text{weak}_s$ with $\{s^1\}$ and $\mathcal{L}^\text{weak}_a$ with $\{(s^1,a^1)\}$ according to Eqn. (9)\;%~\eqref{eqn:weak_min}
  Train $\Phi$ with the objectives in Eqn. (1), (7) and (9)\;%~\eqref{eqn:gans},~\eqref{eqn:dyncon} and~\eqref{eqn:weak_min}
  Sample a batch of state-action pairs $\{s^1\}$ from $\Xi^1$ and $\{s^2\}$ from $\Xi^2$\;
  Train $D_s$ with $\{s^1\}$ and $\{s^2\}$ according to the objective in Eqn. (1)\;%~\eqref{eqn:gans}
  }
  Fix the parameters of $\Phi$ and $D_s$\;
  \For{$i=1\rightarrow N^a$}{
  Sample a batch of state-action pairs and actions: $\{s^1,a^1\}$ and $\{a^1\}$ from $\Xi^1$ and $\{s^2,a^2\}$ and $\{a^2\}$ from $\Xi^2$\;
  Compute $\mathcal{L}^a_\text{adv}$ with $\{a^2\}$, $\{s^1,a^1\}$, $\{a^1\}$ and $\{s^2,a^2\}$ according to Eqn. (2)\;%~\eqref{eqn:gana}
  Sample a batch of state-action pairs $\{s^1,a^1\}$ from $\Xi^1$ and $s^2,a^2\}$ from $\Xi^2$\;
  Compute $\mathcal{L}_\text{m\_dom\_con}$ with $\{s^1,a^1\}$ and $\{s^2,a^2\}$ according to Eqn. (3)\;%~\eqref{eqn:domcon}
  Sample a batch of segments $\{s^1_t,a^1_t,\cdots,s^1_{t+T}\}$ from $\Xi^1$\;
  Compute $\mathcal{L}_\text{m\_dyn\_con}$ with $\{s^1_t,a^1_t,\cdots,s^1_{t+T}\}$ according to Eqn. (7)\;%~\eqref{eqn:dyncon}
  Sample a batch of states $\{s^1\}$ and state-action pairs $\{(s^1,a^1)\}$ from $\Xi^1$\;
  Compute $\mathcal{L}^\text{weak}_s$ with $\{s^1\}$ and $\mathcal{L}^\text{weak}_a$ with $\{(s^1,a^1)\}$ according to Eqn. (9)\;%~\eqref{eqn:weak_min}
  Train $H_1$ and $H_2$ with the objectives in Eqn. (2), (3), (7) and (9)\;%~\eqref{eqn:gana},~\eqref{eqn:domcon},~\eqref{eqn:dyncon} and~\eqref{eqn:weak_min}
  Sample a batch of state-action pairs and actions: $\{s^1,a^1\}$ and $\{a^1\}$ from $\Xi^1$ and $\{s^2,a^2\}$ and $\{a^2\}$ from $\Xi^2$\;
  Train $D_{a^1}$ and $D_{a^2}$ with $\{s^1\}$ and $\{s^2\}$ according to the objective in Eqn. (2)\;%~\eqref{eqn:gana}
  }
  }
  \KwOut{The learned state map $\Phi$ and the learned action maps $H_1$ and $H_2$.}
  \caption{Algorithm for learning the maps}\label{alg:algo2}
\end{algorithm}

\section{Experiment Details}

\subsection{Confidence as Loosely Paired Data}
\begin{table*}[ht]
    \centering
    \begin{tabular}{c|ccccccc}
    0.1 &  2.76  & 294.93 & 796.69 & 1360.38 & 1831.08 & 2344.46 & 3826.94 \\
    0.9 & 2.76  & 294.93 & 796.69 & 1360.38 & 1831.08 & 2344.46 & 3035.05 \\ 
    \end{tabular}
    \caption{Average Return of Demonstrations that are used to generate policy Walker2d.}
    \label{tab:demos_confidence_walker}
\end{table*}

\begin{table*}[ht]
    \centering
    \begin{tabular}{c|ccccccc}
    both &  20.80  & 57.78 & 230.73 & 701.21 & 1546.88 & 2375.35 & 3106.08 \\
    \end{tabular}
    \caption{Average Return of Demonstrations that are used to generate policy Hopper.}
    \label{tab:demos_confidence_hopper}
\end{table*}

{\small
\bibliography{main}
\bibliographystyle{IEEEtran}
}